# A Heuristic Method to Generate Better Initial Population for Evolutionary Methods

E. Khaji[a], A. S. Mohammadi[b]

a,b Graduate Students, Complex Adaptive Systems Group, School of Applied Physics, University of Gothenburg, Sweden.

**Abstract**

Initial population plays an important role in heuristic algorithms such as GA as it help to decrease the time those algorithms need to achieve an acceptable result. Furthermore, it may influence the quality of the final answer given by evolutionary algorithms. In this paper, we shall introduce a heuristic method to generate a target based initial population which possess two mentioned characteristics. The efficiency of the proposed method has been shown by presenting the results of our tests on the benchmarks.

**Introduction**

Evolutionary algorithms (EAs) have been introduced to solve nonlinear complex optimization problems [1],[2] and [3]. Some well-established and commonly used EAs are Genetic Algorithms (GA) [4] , Particle Swarms Optimization [5] and Differential Evolution (DE) [6] and [5]. Each of these method has its own characteristics, strengths and weaknesses; but long computational time is a common drawback for all population-based schemes, specially when the solution space is hard to explore [hamin maghala di]. Many efforts have been already done to accelerate convergence of these methods. Most of these works are focused on introducing or improving crossover and mutation operators, selection mechanisms, and adaptive controlling of parameter settings.[hamin maghala di] Although, population initialization can affect the convergence speed and also the quality of the final solution, there is only a little reported research in this field. Maaranen et al. introduced quasi-random population initialization for genetic algorithms [7]. The presented results showed that their proposed initialization method can improve the quality of final solutions with no noteworthy improvement for convergence speed. On the other hand, generation of quasi-random sequences is more difficult and their advantage vanishes for higher dimensional problems (theoretically for dimensions larger than 12) [8]. Also, Rahmanian et al [Hmian maghala di] presented a novel scheme for population initialization by applying opposition-based learning [9] to make EAs faster.

This paper presents a heuristic method based on the combination of classical and evolutionary methods. We have suggested that solving some simplified equations of a complex function in addition to some random individuals can make a novel initial population which converges to an acceptable answer faster.

Organization of the rest of the paper is as follows: in Section 2, the concept of opposition-based learning is briefly explained. The classical DE is briefly reviewed in Section 3. The proposed algorithm is presented in Section 4. Experimental results are given in Section 5. The results are analysed in Section 6. Finally, the work is concluded in Section 7. All benchmark functions are listed in Appendix.

## Proposed Method

Let X* be a local minimum on n-dimensional function of

$$f(X_n), n \in N.$$

If $\exists\, r \in N$ so that for every point such $B: |B - X^*| < r, f'(B) \neq 0, and\ f(B) \neq \{\}$, the region $f(A)$, where $A$ is a $n-1$ dimensional sphere with the radius of $r$ and $X^*$ as its center, is called a CAVITY. Every function such f can be secreted into a union of cavities which may to be empty while each cavity involves one and only one local minimum.

In order to have an appropriate initial population, generating at least one point in each cavity is clearly fruitful. Although there is no guarantee to generate an initial population which possess such characteristic, our method aims to introduce a heuristic approach to generate initial population which can cover too many cavities.

**The Main Proposed Method:** Suppose $f_n(x)$ as an n-dimensional continuous function on R$^n$. The initial population for approaching this function is the solution of the following equation:

$$F(x_1, x_2, x_3, x_4, x_5, \ldots, x_m, \ldots, x_n) = F(x_1, x_2, x_3, x_4, x_5, \ldots, x_m + \varepsilon, \ldots, x_n). \quad (1)$$

Theoretically, the intersections are $n-1$ dimensional functions for $\varepsilon > 0$. The task is just to find a point in each $n-1$ dimensions. As the task seems too challenging in application since solving the so called equation is not possible all the time, we are suggesting several tricks to tackle the mentioned difficulty.

1. Function Separation: if the function $f$ can be separated as follows:

$$F(x) = f_1(x)f_2(x) + f_3(x)f_4(x) + \cdots$$

then the calculated points from the following equations may have the same quality as the solutions of (1):

$$F_n(x_m) = f_1(x_m + \varepsilon)f_2(x) + f_3(x)f_4(x) + \cdots$$
$$F_n(x_m) = f_1(x)f_2(x_m + \varepsilon) + f_3(x)f_4(x) + \cdots$$
$$F_n(x_m) = f_1(x)f_2(x) + f_3(x_m + \varepsilon)f_4(x) + \cdots$$

2. Dimension Separation: if the function f can be separated as follows:

$$F_n(x) = f_1(x_1) + f_2(x_2) + f_3(x_3) + \cdots,$$

then the calculated points from the following equations may have the same quality as the solutions of (1):

$$F_1(x_1) = f_1(x_1 + \varepsilon)$$

$$F_2(x_2) = f_2(x_2 + \varepsilon)\ldots$$

$$F_n(x_n) = f_n(x_n + \varepsilon)$$

## Experimental Results

In order to compare convergence speed of the suggested method with random population initialization (DEr) some test set with 6 numerical benchmark functions is employed. All selected functions are well-known in the global optimization literature [10,13,15]. The test set includes unimodal as well as highly multimodal minimization problems. The dimensionality of problems varies from 2 to 100 to cover a wide range of problem complexity. The definition, the range of search space, and also the global minimum of each function are given in Appendix.

We compare the convergence speed of our method and DEr by measuring the number of function calls (NFC) which is the most commonly used metric in the literature [14,12]. Smaller NFC means higher convergence speed (for more theoretical information about the convergence properties of evolutionary algorithms the reader is referred to [11]). In all solved problems, $\varepsilon$ is the lowest positive value one can chose. Also, in all solved problems, size of initial population is 100.

1- Sphere Model, $-5.12 < x_i < 5.12$, $min(f_1) = f_1(0,0,\ldots,0) = 0$.

$$\sum_{i=1}^{30} x_i^2 = \sum_{i=1}^{30}(x_i - \varepsilon)^2 \rightarrow \sum_{i=1}^{30}(x_i - \varepsilon)^2 - x_i^2 = 0,$$

$$\sum_{i=1}^{30} \varepsilon^2 - 2\varepsilon x_i = 0 \rightarrow \sum_{i=1}^{30} 2\varepsilon x_i = \varepsilon^2 \rightarrow$$

Since $x_i$'s independently influences $f$, then $min(f) = min(x_1) + min(x_2)\ldots min(x_n)$. Hence:

$$x_i = \frac{\varepsilon}{2}, \quad for\ i = 1,2..,30.[1]$$

| NFC(DE) | NFC(Suggested Method) |
|---------|----------------------|
| 28151   | ***100[2]***         |

2- Axis Parallel: $-5.12 < x_i < 5.12$, $min(f_2) = f_2(0,0,\ldots,0) = 0$.

$$\sum_{i=1}^{30} i x_i^2 = \sum_{i=1}^{30} i(x_i - \varepsilon)^2 \rightarrow \sum_{i=1}^{30} i x_i^2 - i(x_i - \varepsilon)^2 = 0 \rightarrow x_i = \varepsilon/2,\ for\ i = 1,2..,30.$$

| NFC(DE) | NFC(Suggested Method) |
|---------|----------------------|
| 37555   | ***100***            |

___________________________________________________________________

1: It is obvious that the less the value of $\varepsilon$, the more the accuracy of the result. In experiments, $\varepsilon = 0$.
2: NFC=100 means that the final answer is among the initial population.

3- Rosenbrock Function: $-2 < x_i < 2$, $min(f_3) = f_3(1,1,..,1) = 0$.

$$\sum_{i=1}^{30}[100(x_{i+1} - x_i^2)^2 - (1 - x_i)^2] = \sum_{i=1}^{30}[100(x_{i+1} - \varepsilon - x_i^2)^2 - (1 - x_i)^2],$$

$$\varepsilon - 2x_{i+1} + 2x_i^2 = 0 \rightarrow x_i^2 = (x_{i+1} - \varepsilon)/2. \quad (1)$$

According to (1) and the main function:

$$x_i = 1 - \sqrt{-100\varepsilon^2/4}. \quad for\ i = 1,2..,30.$$

| NFC(DE) | NFC(Suggested Method) |
|---|---|
| 295255 | ***100*** |

4- Rastrigin Function: $-5.12 < x_i < 5.12$, $min(f_4) = f_4(0,0,...,0) = 0$.

$$100 + \sum_{i=1}^{10}(x_i^2 - 10\cos(2\pi x_i)) = 100 + \sum_{i=1}^{10}((x_i - \varepsilon)^2 - 10\cos(2\pi x_i)),$$

$$\sum_{i=1}^{10}((x_i - \varepsilon)^2 - x_i^2) = 0 \rightarrow x_i = \frac{\varepsilon}{2}, for\ i = 1,2..,10.$$

For cosine function we have:

$$100 + \sum_{i=1}^{10}(x_i^2 - 10\cos(2\pi x_i)) = 100 + \sum_{i=1}^{10}(x_i^2 - 10\cos(2\pi x_i - \varepsilon)),$$

$$\sum_{i=1}^{10}(10\cos(2\pi x_i)) - \sum_{i=1}^{10}(10\cos(2\pi x_i - \varepsilon)) = 0,$$

$$\sum_{i=1}^{10}(\cos(2\pi x_i) - \cos(2\pi x_i - \varepsilon)) = 0,$$

$$\sum_{i=1}^{10} -2\sin((4\pi x_i - \varepsilon)/2)\sin(\varepsilon/2) = 0.$$

Hence,

$$\sin(4\pi x_i - \varepsilon) = 0 \rightarrow 4\pi x_i - \varepsilon = \pm k\pi \rightarrow x_i = \frac{\pm k\pi + \varepsilon}{4\pi}.$$

| NFC(DE) | NFC(Suggested Method) |
|---------|----------------------|
| 383377  | ***100***            |

5- Brainins' function $-5 < x_1 < 10, 0 < x_2 < 15, min(f_5) = f_5(-\pi, 12.275)/(-\pi, 2.275)/(9.42478, 2.475) = 0.3979$.

$$a(x_2 - bx_1^2 + cx_1 - d)^2 + e(1-f)\cos(x_1) + e$$
$$= a(x_2 - bx_1^2 + cx_1 - d - \varepsilon)^2 + e(1-f)\cos(x_1) + e,$$

$$x_2 = \frac{-\varepsilon + 2(-bx_1^2 + cx_1 - d)}{2},$$

$$x_1 = \frac{b\varepsilon + c}{2b}.$$

$$a(x_2 - bx_1^2 + cx_1 - d)^2 + e(1-f)\cos(x_1) + e$$
$$= a(x_2 - bx_1^2 + cx_1 - d)^2 + e(1-f)\cos(x_1 - \varepsilon) + e,$$

$$\cos(x_1) - \cos(x_1 - \varepsilon) = 0,$$

$$-2\sin((2x_1 - \varepsilon)/2)\sin(\varepsilon/2) = 0,$$

$$x_1 = \pm k\pi + \varepsilon/2,$$

$$x_2 = \frac{-\varepsilon + 2(-b\pi^2 + c\pi - d)}{2}.$$

| NFC(DE) | NFC(Suggested Method) |
|---------|----------------------|
| 4918    | ***100***            |

6- Michalewicz function, $0 < x_i < \pi, min(f_{6_{n=5}}) = -4.687658$.

$$-\sum_{i=1}^{5} \sin(x_i) \sin\left(\frac{ix_i^2}{\pi}\right)^{20} = -\sum_{i=1}^{5} \sin(x_i - \varepsilon) \sin\left(\frac{ix_i^2}{\pi}\right)^{20},$$

For $sin(x_i)$ function we have:

$$\sum_{i=1}^{5} \sin(x_i) = \sum_{i=1}^{5} \sin(x_i - \varepsilon),$$

$$\sum_{i=1}^{5} 2\cos((2x_i - \varepsilon)/2) \sin(\varepsilon/2) = 0,$$

$$\cos((2x_i - \varepsilon)/2) = 0,$$

$$\frac{2x_i - \varepsilon}{2} = k\pi \pm \frac{\pi}{2},$$

$$x_i = k\pi \pm \frac{\pi}{2} + \frac{\varepsilon}{2}.$$

| NFC(DE) | NFC(Suggested Method) |
|---------|----------------------|
| 3132    | **100**              |

Figure 1: The performance of complete random population, semi-random population, and selected initial population on the Michalewicz function after 40 runs for each sort of initial population.

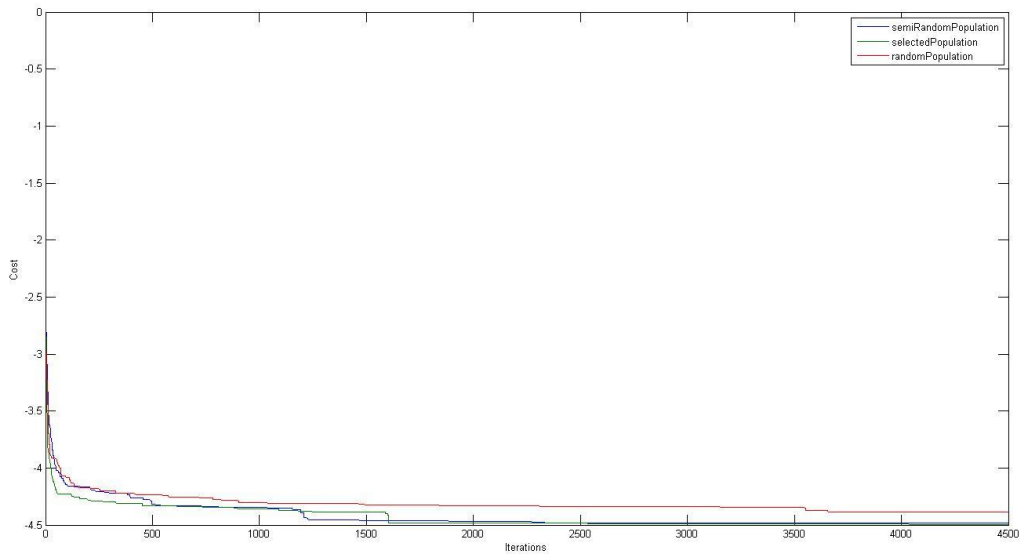

| NFC(DE) | NFC(Suggested Method) |
|---------|----------------------|
| 4500    | **4136**             |

7- Matyas function, $-10 < x_i < 10$, $min(f_7) = f_7(0,0) = 0$.

$$0.26((x_1 - \varepsilon)^2 + x_2^2) - 0.48 x_1 x_2 = 0.26(x_1^2 + x_2^2) - 0.48 x_1 x_2 \quad (1)$$

$$x_1 = -\frac{0.48}{0.52} x_2 + \frac{0.26\varepsilon}{0.52} \quad (2)$$

Inserting (2) in (1), we have:

$$x_2 = \frac{\varepsilon}{2}.$$

| NFC(DE) | NFC(Suggested Method) |
|---------|----------------------|
| 3132    | **100**              |

# Conclusion

Our heuristic method has been applied into some famous benchmarks. The main idea of the method was to combine classical and arbitrary nature of the existed methods in order to generate more proper initial population for heuristic methods. Results show that in many continuous functions, the final answer can be obtained in selected population suggested by our method. In some cases such as f(6), the semi-random population work more efficiently within the time which indicates that the obtained answers with combination of random numbers in the initial population led to the best result faster than the case where the initial population includes random numbers or selected answers. Hence, it can be concluded from the experiments that with the same size of initial population, a semi-random population including selected results of solved equations and random numbers can outperform those include random numbers.